\documentclass[conference]{IEEEtran}

\usepackage{graphicx}
\usepackage{amsmath,amssymb}
\usepackage{booktabs}
\usepackage{array}
\usepackage{multirow}
\usepackage{cite}
\usepackage{url}
\usepackage{xcolor}
\usepackage{xspace}
\usepackage[hidelinks]{hyperref}
\usepackage[caption=false,font=footnotesize]{subfig}
\usepackage{xurl}

\newcommand{\method}{AmodalDINO\xspace}
\newcommand{\eg}{e.g.,\ }
\newcommand{\ie}{i.e.,\ }
\newcommand{\etal}{et al.\ }
\newcommand{\brk}{\discretionary{}{}{}}

\title{Foreseeing the Invisible:\\Amodal Reconstruction of Leaf Fossil Images}

\author{
\IEEEauthorblockN{Liuxiang Yue, Ailin Zhang, Ziyue Zhao, Yikun Duan}
\IEEEauthorblockA{
Global College\\
Shanghai Jiao Tong University\\
Shanghai, China\\
yueliuxiang@sjtu.edu.cn
}
}

\begin{document}

\maketitle

\begin{abstract}
Fossil leaves are rarely preserved whole---sedimentary rock hides, breaks, and erodes the
lamina, yet paleobotany depends on the \emph{complete} shape and outline of the leaf. We
cast the recovery of the missing tissue as amodal reconstruction and present
\method, a multi-head dense-prediction model that predicts four masks from a single RGB image:
visible leaf, amodal complete leaf, amodal main vein, and fine
veins. Unlike essentially all prior amodal work, \method is given no visible
mask. It predicts the visible and amodal regions jointly, so it needs no upstream
instance segmenter at runtime. Two simple but effective changes adapt the model
to the amodal segmentation task: fully fine-tune a
DINOv3 ViT-L/16 at a small learning rate instead of freezing it, and attach
auxiliary venation heads alongside the leaf heads. These two changes enable the
model to learn the structural shape prior of leaves. Trained only on synthetic leaf fossil images,
\method reaches $95.0\%$ Dice / $90.5\%$ IoU on the validation set and transfers well
to real fossil specimens. Stripped to two heads, the same recipe can run on two benchmark datasets,
reaching $85.05$ full mIoU / $66.65$ occluded mIoU on KINS and $80.90$ / $38.15$ on COCOA-cls.
The model is also practical: by quantizing
to 4-bit weights, it runs entirely offline in a browser, matching the original model with an IoU of 0.910.
We also add ruler-based calibration to estimate surface area, and a generative visualization
of living leaves on local devices.
\end{abstract}


\section{Introduction}
\label{sec:introduction}

A fossil leaf is a partial observation of an object that no longer exists. Rock
covers part of the blade, cracks split it, and weathering removes the margins, so
the remaining part is almost never the whole leaf. Paleobotanists, however,
reason about the \emph{complete} organ: the overall silhouette and the venation
pattern carry the diagnostic signal for identifying a species and for
reconstructing past climates from leaf shape~\cite{wilf2016fossilleaf}. Standard
semantic and instance segmentation only label the visible part, which leaves the
missing tissue unrecovered.

We therefore target \emph{amodal} reconstruction: given a leaf-fossil image, we
predict not only the visible leaf but its full spatial extent and its veins,
effectively imagining the parts hidden by stone. This is harder than road-scene
or everyday-object amodal segmentation~\cite{qi2019kins,zhu2017cocoa} for the following
reasons. First, fossil leaves have no rigid category template to fall back on.
Unlike a car or a pedestrian, every specimen is a different organic shape, so the
network cannot memorize a canonical silhouette. Second, a lack of real fossil images
forces a synthetic training dataset and a domain gap to real specimens. Third, as veins are thin,
such low-contrast structures are easily confused with cracks and mineral seams in
the surrounding rock matrix.

\begin{figure}[t]
    \centering
    \includegraphics[width=\columnwidth]{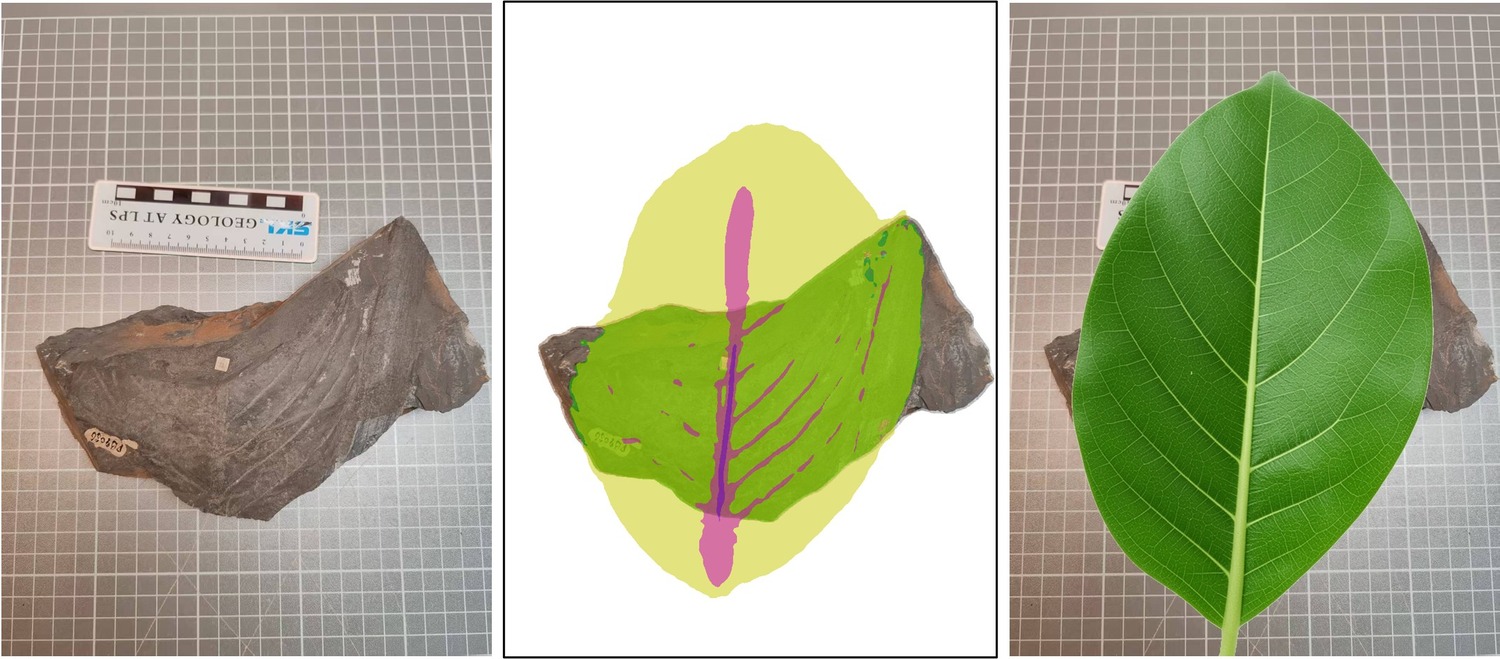}
    \caption{\method on a real fossil specimen it never saw during training. Left:
    the photograph, in which roughly half of the blade is buried in rock. Middle:
    the four predicted masks---visible tissue (green), completed amodal leaf
    (pale yellow), primary vein (magenta), detail veins (purple). Right: an
    optional generative rendering of a living leaf.
    The model extrapolates the blade well beyond the exposed lamina and recovers
    a coherent midrib.}
    \label{fig:teaser}
\end{figure}

\noindent\textbf{No visible mask.} A fourth difference concerns what the model is
allowed to see, and it separates our setting from the literature more sharply
than the fossil domain does. Almost all the other amodal methods are handed the
visible mask as input, either as ground truth (the ``Oracle''
protocol) or from an upstream segmenter (the ``Standard'' protocol),
to grow that mask outwards. The assumption is reasonable for cars and chairs, but
does not survive contact with a fossil: there is no good detector for ``leaf
embedded in rock,'' and the boundary between lamina and background is ambiguous.
\method therefore consumes no visible mask at
all: it takes RGB and emits the visible mask as one of its four outputs. The
network does accept one optional extra input, a coarse region-of-interest (ROI)
hint that lets a user point at a specimen in the interactive demo, but it is not
part of the method. Every number we report is measured with default neutral ROI.

Our approach starts from a strong self-supervised backbone, DINOv3
\cite{simeoni2025dinov3}, but departs from its recommended usage in two ways that
turn out to be the crux of the problem.

\noindent\textbf{Fully fine-tune, do not freeze.} The official recipe for pre-trained
DINOv3 is to treat it as a \emph{frozen} feature extractor with a
trainable decoder. But for amodal tasks, this leaves most of the work undone. A frozen
backbone only extrapolates a small, limited envelope that ignores
the geometry of the specimen in front of it. Unfreezing every layer at a small learning rate
changes the character of the prediction within a few epochs: margins snap to the
specimen, lobes and tips appear, and the model still generalizes, because
a small step size adapts the features rather than overwriting them.

\noindent\textbf{Let structure supervise shape.} The second choice is which heads
hang off the shared trunk, and it acts on the same axis. An amodal head trained
alone can complete slightly damaged fossils, but it often predicts a
rounded, low-frequency blob. Adding a visible head gives the model an explicit place to say
where observation ends and inference begins. Adding an amodal vein head
supplies the shape prior, because a coherent venation can only be drawn on a
coherent lamina, so supervising the midrib forces the trunk to represent lobes and
sinuses. A final detail-vein head, although noisy in isolation, makes the primary-vein head
lock onto the true midrib and reject stone cracks.

Around this model we build a complete, deployable system: a synthetic
data-generation pipeline based on the NMNS Cleared Leaf Database, an interactive
browser demo that runs a 4-bit quantized model with ONNX Runtime Web, and a local
GPU pipeline that generates realistic leaf visualizations.

\noindent\textbf{Contributions.}
\begin{itemize}
    \item We formulate the amodal reconstruction of leaf fossils: joint recovery
    of the complete leaf silhouette and its venation under stone occlusion from a
    single RGB image, and build \method, an end-to-end model for it that
    requires no visible-mask input.
    \item We identify two design choices that make completion work with scarce
    labels: full fine-tuning of a self-supervised DINOv3 backbone rather than
    freezing it, and auxiliary venation heads that supply a structural prior. They
    significantly improve the fidelity of the completed shape.
    \item We give a fully specified, reproducible recipe: the official DPT trunk,
    four independent heads, a completion-oriented multi-head loss, a two-rate
    schedule that DINOv3 turns out to require, and report per-head results on
    synthetic fossils together with KINS and COCOA-cls numbers obtained without
    any visible-mask input.
    \item We demonstrate practical deployment: a quantized in-browser model and a local
    GPU pipeline, featuring a ruler-based scale, no-ground-truth
    quality scoring, and optional generative leaf revival.
\end{itemize}

\section{Related Work}
\label{sec:related_work}

\textbf{Amodal instance segmentation.} KINS~\cite{qi2019kins} and
COCOA~\cite{zhu2017cocoa} added amodal masks to KITTI and COCO and established
the full / occluded mIoU protocol we borrow for baselines. A first line of work
predicts the occluded mask from an explicit visible cue: PCNet~\cite{zhan2020pcnet}
de-occludes a scene in a self-supervised manner; VRSP~\cite{xiao2021vrsp} and
C2F-Seg~\cite{gao2023c2fseg} regularize completion with a learned shape prior;
AISFormer~\cite{tran2022aisformer} reasons over RoI features with a transformer;
and the recent GRASP~\cite{zhang2026grasp} combines learnable shape prototypes
with a gate driven by the signed distance field of the visible mask. A second
line adapts foundation models: PLUG~\cite{liu2024plug} attaches parallel LoRA
branches to SAM~\cite{kirillov2023sam} and prompts it with the visible bounding
box, and Amodal SAM~\cite{zhang2026amodalsam} adds a spatial-completion adapter,
currently the strongest reported numbers on KINS and COCOA. A third line
sidesteps masks entirely and \emph{synthesizes} the whole object with a diffusion
prior, as in pix2gestalt~\cite{ozguroglu2024pix2gestalt} and open-world
appearance completion~\cite{ao2025openworld}, while Zhan~\etal~\cite{zhan2024wild}
obtain authentic amodal ground truth from 3D scans rather than simulating
occlusion.

Two properties are common to nearly all of this work but do not hold in
our setting: it targets rigid, common object categories with a strong shape
prior, and it consumes a ground-truth or predicted visible mask as input.
Notably, even GRASP, which like us builds on
a self-supervised DINO backbone, keeps that backbone \emph{frozen} and puts its
capacity into a prototype module driven by the given visible mask. We instead
complete organic, category-free leaf silhouettes \emph{together with} their
internal venation, and treat the visible mask as an output rather than an input.

\textbf{Self-supervised transformers for dense prediction.} DINOv2 and DINOv3
\cite{oquab2024dinov2,simeoni2025dinov3} learn transferable visual features
without labels and are usually paired with a lightweight dense decoder such as
DPT~\cite{ranftl2021dpt}, originally proposed for monocular depth. The standard
protocol keeps the backbone frozen and trains only the decoder, on the reasoning
that the pretrained features are already close to optimal and that fine-tuning
risks destroying them. Our contribution is not the backbone itself but how it is
adapted: we fully fine-tune it, repurpose the DPT trunk to drive four amodal
heads, and show that the frozen protocol is precisely what prevents
extrapolation into occluded regions.

\textbf{Computational paleobotany and leaf analysis.} Prior computer-vision work
on leaves focuses on classifying or measuring \emph{intact} cleared specimens,
\eg the leaf-code study of Wilf~\etal~\cite{wilf2016fossilleaf}, which recognizes
plant families from clean leaf images. Recovering the shape and venation of a
\emph{broken, stone-occluded} fossil is unaddressed, and it is
the gap this work targets. The completion aspect is loosely related to
structure-aware image inpainting such as SAIN~\cite{wang2026sain}; unlike RGB
inpainting, we predict masks and explicit venation rather than pixels, and we are
judged on region overlap rather than perceptual realism.

\textbf{On-device and in-browser inference.} Deploying a foundation-scale model
to a laptop or a browser requires quantization and a portable runtime. We rely on
weight-only 4-bit quantization in the spirit of GPTQ~\cite{frantar2023gptq} and
on ONNX Runtime Web~\cite{onnxruntime} through a WebGPU backend,
together with a real-time open-vocabulary detector, YOLO26-seg~\cite{wang2025yoloe},
and a compact promptable segmenter whose text encoder is
MobileCLIP~\cite{vasu2024mobileclip}.

\section{Method}
\label{sec:method}

\begin{figure*}[t]
    \centering
    \includegraphics[width=0.90\textwidth]{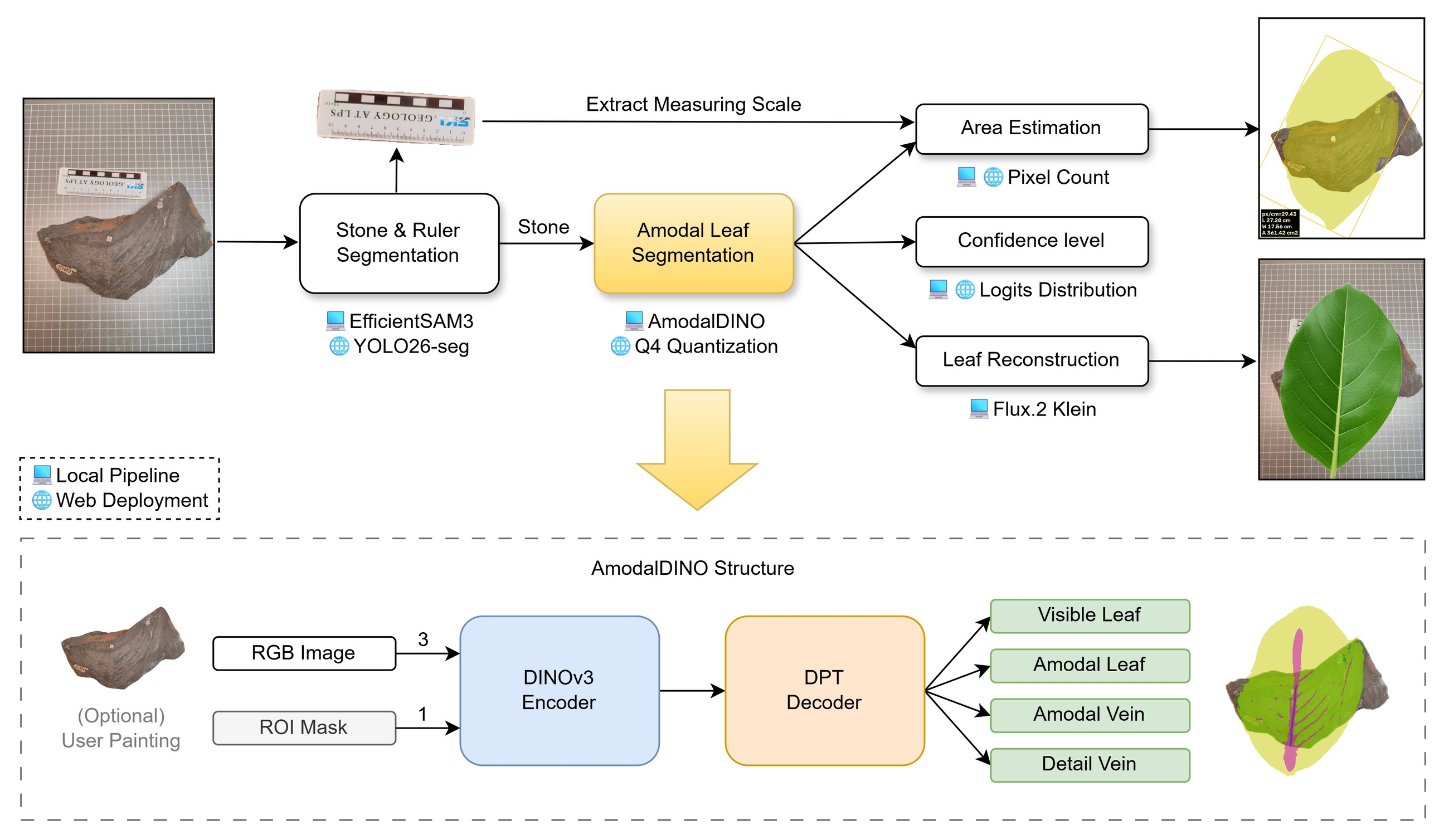}
    \caption{Top: the deployed system. A photograph of a specimen is segmented
    into stone and ruler, the stone cut-out is passed to \method, and the four
    predicted masks drive a pixel-count area estimate, a no-ground-truth
    confidence read-out, and an optional generative leaf rendering.
    Bottom: \method itself---RGB (3 channels) and an
    optional ROI mask (1 channel) enter a fully fine-tuned DINOv3 ViT-L/16
    encoder, a DPT decoder fuses four intermediate token blocks, and four
    independent convolutional heads emit the visible leaf, the amodal leaf, the
    amodal vein, and the detail vein.}
    \label{fig:arch}
\end{figure*}

\subsection{Problem Formulation}
Given an RGB image $\mathbf{I}\in\mathbb{R}^{3\times H\times W}$ and an optional
region-of-interest (ROI) hint $\mathbf{R}\in\{0.5,1\}^{1\times H\times W}$, we
form a 4-channel input $\mathbf{x}=[\mathbf{I};\mathbf{R}]$ and predict a
4-channel logit map
$\mathbf{y}=f_\theta(\mathbf{x})\in\mathbb{R}^{4\times H\times W}$. The channels
are, in fixed order, the visible leaf $M_v$, the complete amodal leaf $M_a$, the
amodal primary vein $M_{av}$, and the fine detail vein $M_{dv}$; each mask is
read off by thresholding, $\hat{M}_c=\sigma(\mathbf{y}_c)>0.5$. The quantity of
scientific interest is the occluded region $M_a\setminus M_v$, \ie the tissue the
camera never saw, which the model must hallucinate consistently with the visible
evidence.

\noindent\textbf{The role of the ROI channel.} The hint exists for the
interactive browser demo, where a user may want to point at one specimen or nudge
a completion by hand. It is not part of the core recipe: batch inference passes a
neutral hint, all fossil numbers in Section~\ref{sec:fossil-results} are measured
that way, and the public-benchmark variant of Section~\ref{sec:baselines} drops
the ROI channel altogether. The one design decision worth stating here is that we
encode a positive hint as $1$ and ``no information'' as $0.5$, rather than the
obvious $\{0,1\}$, and normalize the channel with mean and standard deviation
$0.5$. A neutral hint then maps exactly to zero, contributes nothing to the patch
embedding, and leaves the network in pure-RGB mode: a hint only ever \emph{adds}
evidence, and its absence is not itself a signal.
Appendix~\ref{app:roi} covers the encoding, the sampling policy, and the
corruption schedule that keeps the hint from being treated as a leaked
silhouette.

\subsection{Architecture}
Figure~\ref{fig:arch} shows the network in the context of the full system. It has
three parts: an ROI-aware DINOv3 backbone, a DPT trunk, and four independent
convolutional heads.

\noindent\textbf{Backbone.} We use a DINOv3 ViT-L/16 backbone (embedding
dimension $1024$, patch size $16$, 24 transformer blocks) initialized from the
public LVD-1689M weights~\cite{simeoni2025dinov3}, with RGB normalized by
ImageNet statistics. Inputs are reflect-padded to a multiple of the patch size
and the logits are cropped back after decoding. To admit the extra ROI channel
without discarding the pretrained stem, we widen the patch-embedding convolution
from 3 to 4 input channels, copying the RGB filters and initializing the fourth
as their per-position mean; together with the neutral-hint encoding above, the
network starts from exactly its pretrained behavior.

\noindent\textbf{DPT trunk.} We tap the token sequence at four evenly spaced
transformer blocks---indices $\{4,11,17,23\}$ for the 24-block ViT-L---and fuse
them with the official DINOv3 DPT trunk~\cite{ranftl2021dpt}
(\texttt{readout=project}, trunk width $256$, per-stage post-processing channels
$[128,256,512,768]$). The trunk reassembles the four token maps to
$\{4{\times},2{\times},1{\times},\tfrac{1}{2}{\times}\}$ of the patch grid, fuses
them top-down, and emits a single $256$-channel dense feature map at half the
input resolution. We keep the decoder light so that representational
power comes from the fine-tuned backbone rather than from a heavy task-specific
head; this is consistent with our finding that adaptation must happen in the
backbone.

\noindent\textbf{Independent heads.} Four independent $3\times3$ convolutions
($256\to1$) map the shared trunk features to the four masks. Keeping the heads
separate rather than using a single 4-channel convolution lets each task
specialize while still sharing the trunk, and it makes the progressive-stacking
analysis of Section~\ref{sec:ablation} clean, since heads can be added or removed
without touching the rest of the network. The stacked logits are bilinearly
upsampled to the padded resolution and cropped back to $H\times W$.

Crucially, the backbone is \emph{fully unfrozen}. As Section~\ref{sec:ablation}
shows, the recommended frozen backbone with a trainable decoder still completes,
but only into a generic rounded envelope, whereas full
fine-tuning at a small learning rate ($10^{-5}$ for the backbone, $10^{-4}$ for
the trunk and heads) recovers specimen-specific margins within a few epochs while
preserving the generalization of the pretrained features.

\subsection{Completion-Oriented Multi-Head Loss}
\label{sec:loss}
The loss is built around a single tension. The completion heads must grow the
mask into unobserved territory, which rewards recall, without spilling past the
true leaf margin or painting stone as tissue, which would destroy precision. We
resolve it with a boundary-weighted cross-entropy, region-specific reweighting,
asymmetric overlap terms, and soft structural constraints.

\noindent\textbf{Boundary-weighted BCE.} Every head uses a binary cross-entropy
whose per-pixel weight emphasizes the target interior and, more strongly, its
edge:
\begin{equation}
\label{eq:bw}
w = 1 + 2\,T + 4\,\partial T,
\end{equation}
where $T$ is the target mask and $\partial T$ its one-pixel morphological
boundary (a $3{\times}3$ dilation minus erosion). This concentrates gradient on
thin structures and margins, which matters for both leaf outlines and veins. Note
that $w$ here is a per-pixel \emph{weight map} inside the cross-entropy, not a
loss term.

\noindent\textbf{Head-specific overlap terms.} The visible and detail-vein heads
add a soft Dice term~\cite{milletari2016vnet} to the weighted BCE. The two
completion heads instead pair the BCE with a soft Tversky
loss~\cite{salehi2017tversky},
\begin{equation}
\label{eq:tversky}
\mathcal{L}_{\text{Tv}} = 1 - \frac{TP + \epsilon}
{TP + \beta_{\text{FP}}\,FP + \beta_{\text{FN}}\,FN + \epsilon},
\end{equation}
with $\beta_{\text{FP}}{=}0.65$ and $\beta_{\text{FN}}{=}0.35$, so that
over-growth into stone is penalized more than a missed pixel. Two further
penalties keep the hallucinated region tight: an \emph{area-overflow} term that
grows with the squared relative excess of predicted over ground-truth area, and a
\emph{false-positive pixel} term on the mean predicted probability outside the
target.

\noindent\textbf{Completion reweighting.} A model can score well on Dice by
simply reproducing the visible mask, because the occluded region is a minority of
pixels. To counter this, the amodal BCE is up-weighted precisely where the camera
saw nothing:
\begin{equation}
\label{eq:reweight}
w_a = 1 + \lambda_c\underbrace{[\,M_a\wedge\neg M_v\,]}_{\text{completion region}},
\qquad \lambda_c = 2 ,
\end{equation}
and the amodal-vein head is reweighted analogously on its own completion region.
Equation~\eqref{eq:reweight} tells the model that the pixels it will be judged on
are exactly the ones it cannot see.

\noindent\textbf{Soft structural constraints.} Three containment terms inject
leaf anatomy as differentiable priors: the visible mask should lie inside the
amodal leaf, and both vein maps should lie inside the leaf. Each is the mean of
the product of the inner probability and the complement of the outer probability,
with the outer term detached so that the constraint pushes only the inner head
and cannot be satisfied by inflating the outer one.

The total objective sums the four head losses and the three containment terms,
\begin{equation}
\label{eq:total}
\mathcal{L} = \sum_{c} \alpha_c \mathcal{L}_c \;+\; \sum_{k} \gamma_k \mathcal{C}_k ,
\end{equation}
with head weights $\alpha = (1, 1, 1, 0.4)$ and containment
weights $\gamma = (0.5, 0.5, 0.3)$. The detail-vein head is
down-weighted because the quality of ground truth is relatively low.
Every coefficient is listed in Appendix~\ref{app:implementation}.

\section{Synthetic Data Generation}
\label{sec:data}

Real fossils with pixel-level amodal and vein labels do not exist at the scale that is
needed to train a dense-prediction transformer, so we synthesize a larger dataset that
mimics broken, stone-occluded leaves while giving exact ground truth for every
head.

\noindent\textbf{Source specimens.} We start from the NMNS Cleared Leaf Database,
$4{,}041$ high-resolution cleared-leaf photographs in which the full lamina and
venation are visible, and hand-pick $160$ complete leaves among distinct species and
leaf architectures.
These selected leaves are segmented with SAM~3~\cite{carion2025sam3} to isolate the lamina
and give a complete (amodal) silhouette, and their venation is traced by OpenCV followed by manual
correction. This is the only step in the pipeline that costs human effort, and it
is why the count is $160$ rather than $4{,}041$. Deliberately choosing a small, diverse, carefully labelled seed set
over a large noisy one turned out to matter more than we expected, because a
single mislabelled midrib is replicated into every render derived from that leaf.
Everything downstream can be mass produced as follows.

\noindent\textbf{Compositing.} Each leaf is imported into the 3D software \emph{Blender} and projected
onto a randomized stone model with distinct pose, scale, lighting, camera
parameters, etc., under deterministic seeds. Noise textures on the material
simulate the wear and fade of leaf tissues. Because the scene is fully
controlled, every render emits the fossil RGB together with five masks: visible tissue, complete
amodal leaf, primary vein, detail vein, and stone. The renderer produces roughly
five frames per second at $1024\times1024$, so the dataset can be easily expanded. Figure~\ref{fig:blender} in
Appendix~\ref{app:data} shows the scene.

\begin{figure}[t]
    \centering
    \includegraphics[width=0.9\columnwidth]{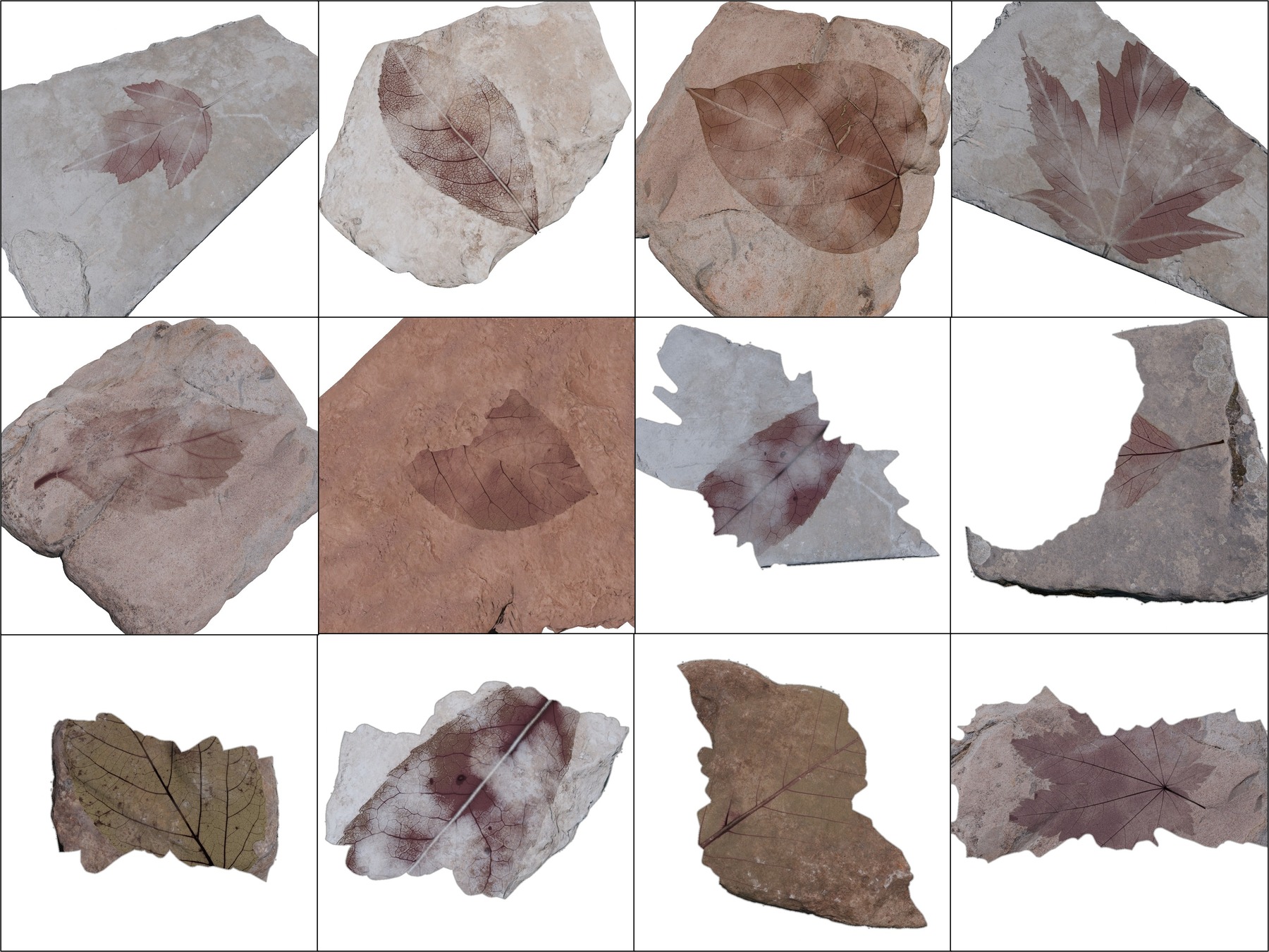}
    \caption{Synthetic leaf fossil images used for training. Pose, scale,
    lighting, camera, stone are randomized in Blender,
    and each rendered scene is then expanded into damaged variants with random
    breakage masks by OpenCV.}
    \label{fig:synthetic}
\end{figure}

\noindent\textbf{Breakage and dataset.} Blender gives us $2{,}200$ intact
composites. Damage is then applied as a cheap post-process rather than in the
renderer: for each composite we draw four random breakage masks in OpenCV, including
ellipses and straight lines with edge distortion, which produces four distinct broken variants per scene and lets us
multiply the dataset without re-rendering. The result is $11{,}000$ images, split into
$10{,}000$ train / $1{,}000$ val images (Fig.~\ref{fig:synthetic}), plus a test
set that contains $9$ photographs of real fossil
specimens (Section~\ref{sec:realfossils}). The model therefore only sees
synthetic fossils during training, and the quality of this dataset matters a lot for
generalizing to real leaf fossils.

The amodal, vein, and detail-vein masks are shared across the five variants of a
scene, since breaking a fossil does not change what the leaf originally was; only
the visible and stone masks are per-variant. Masks are consequently paired to
images by exact basename or, failing that, by stripping successive
\texttt{prefix\_} tokens (\eg
\texttt{broken2\_0001.jpg}$\rightarrow$\texttt{0001.jpg}), which lets the damaged
renders keep descriptive filenames while sharing one set of targets.

\noindent\textbf{Augmentation.} Training applies horizontal and vertical flips
($p{=}0.5$ each), a random multiple of $90^{\circ}$, and isotropic downscaling by
$s\sim\mathcal{U}(0.5,1)$ about a random image point, padding the canvas with
white rather than cropping so that specimens of very different apparent size are
seen at a fixed resolution. Color jitter is deliberately asymmetric in
brightness (a factor in $[0.75,1.65]$) because fossils photograph dark far more
often than bright, with milder contrast, saturation, and hue jitter.

\noindent\textbf{ROI sampling.} The hint is drawn uniformly per sample from
$\{$stone, visible, amodal, amodal-vein, neutral$\}$ and is then heavily
corrupted by random dilation, erosion, and elastic warping, so that the model
learns to treat it as a suggestion rather than a silhouette to trace. Without the
corruption the model latches onto the ROI boundary and collapses the moment a
user draws a sloppy stroke. Because there is a ``neutral'' option, the
same weights also work with no hint at all, which is how every fossil number we
report is measured. Appendix~\ref{app:roi} gives the schedule.

\section{Experiments}
\label{sec:experiments}

\subsection{Implementation Details}
We fine-tune all parameters with AdamW~\cite{loshchilov2019adamw} (weight decay
$10^{-4}$) using two learning rates: $10^{-4}$ for the DPT trunk and heads and
$10^{-5}$ for the backbone, both cosine-annealed to $10^{-6}$ over $40$ epochs.
Training runs at $448\times448$ with batch size $64$ in bfloat16 autocast,
channels-last memory format, and gradient clipping at norm $1.0$, on a single
NVIDIA RTX PRO 6000 (Blackwell); the full schedule takes about two hours. The
$11{,}000$ decoded images are cached in RAM once so that augmentation, not I/O,
is the bottleneck. We select the checkpoint by best validation amodal Dice, at
epoch $37$. Appendix~\ref{app:implementation} lists every coefficient.

\noindent\textbf{DINOv3 is delicate.} The two learning rates are not a rounded
guess; they are the narrowest part of the recipe. Training the backbone at
$10^{-4}$, a value that is unremarkable for the trunk and heads, will lead to instability.
At $10^{-5}$ the model learns leaf shape stably, and the fine veins only
appear once cosine annealing takes the rate below $10^{-5}$, late in the schedule.
The ordering is consistent with what the two tasks demand: coarse silhouette
completion is a large, low-frequency change to the representation, while
distinguishing a secondary vein from a hairline crack in the rock matrix is a small,
high-frequency one that a larger step size simply walks over. Anyone reproducing
this should expect the vein heads to remain weak in early epochs.

This sensitivity is specific to DINOv3 rather than to the task. Before settling on
it we tested other backbones such as PVT-v2~\cite{wang2022pvtv2} and
EVA02~\cite{fang2024eva02}, both of which tolerate a flat $10^{-4}$ across the
whole network and converge without staging. However, they cannot generalize well to
real fossils, which pushed us to DINOv3 and the two-rate
schedule. Our $448\times448$ working resolution is a legacy of that search: EVA02
does not accept $512\times512$, so we settled on $448$
and carried the number over to DINOv3.

\subsection{Metrics}
We report per-head Dice (F1), IoU, precision, and recall at threshold $0.5$;
higher is better.

For the public-baseline experiments introduced in Section~\ref{sec:baselines},
we use the standard amodal protocol: full mIoU (mean IoU between predicted and ground-truth
amodal masks) and occluded mIoU,
\begin{equation}
\label{eq:occiou}
\text{occ-IoU}=\text{IoU}\!\left(\hat{M}_a\setminus M_v^{\text{gt}},\;
M_a^{\text{gt}}\setminus M_v^{\text{gt}}\right),
\end{equation}
averaged over occluded instances only. Note that $M_v^{\text{gt}}$ enters
Eq.~\eqref{eq:occiou} only as an evaluation device; it is never given to our
network.

\subsection{Leaf-Fossil Results}
\label{sec:fossil-results}
Table~\ref{tab:fossil} reports \method on the held-out synthetic validation
split. The amodal leaf head (the primary target) reaches $95.0\%$ Dice and
$90.5\%$ IoU: the model reconstructs the complete silhouette, including tissue
hidden by stone, not merely the visible lamina. The visible head is
near-saturated ($98.5\%$ Dice), as expected for observed content, and the gap of
$8.0$ IoU points between the two heads is a fair measure of how much of the task
is genuine hallucination rather than recognition.

\begin{figure}[t]
    \centering
    \includegraphics[width=\columnwidth]{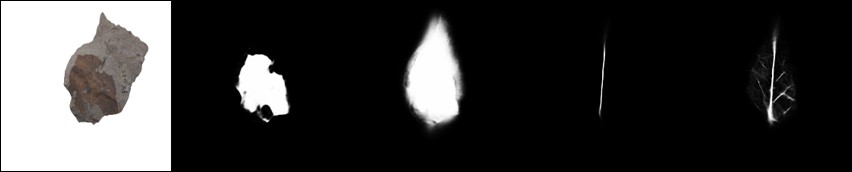}
    \caption{Soft outputs of the four heads for one specimen (left to right:
    input RGB, visible, amodal leaf, amodal vein, detail vein). The amodal head
    grows a plausible blade past the exposed tissue; the vein heads recover the
    midrib while the detail head remains noisier and more diffuse.}
    \label{fig:panels}
\end{figure}

Vein heads are much harder. Veins are one to two pixels wide at the working
resolution and locally indistinguishable from cracks, so Dice drops to $65.4\%$
for the primary vein and $55.3\%$ for the fine veins, even though qualitatively
(Fig.~\ref{fig:teaser}, Fig.~\ref{fig:panels}) the midrib is recovered cleanly.
The large precision--recall asymmetry on the primary vein ($77.5$ vs.\ $56.5$) is
not an accident: the Tversky and overflow penalties of Section~\ref{sec:loss}
were tuned to buy precision with recall, because a conservative vein that a
paleobotanist can trust is more useful than a high-recall one that traces every
fracture in the rock matrix.

\begin{table}[t]
    \centering
    \caption{Per-head results of \method on the synthetic leaf-fossil validation
    split (threshold $0.5$, all values \%). The amodal leaf is the primary
    target.}
    \label{tab:fossil}
    \begin{tabular}{lcccc}
        \toprule
        Head & Dice $\uparrow$ & IoU $\uparrow$ & Prec.\ $\uparrow$ & Rec.\ $\uparrow$ \\
        \midrule
        Visible leaf          & \textbf{98.5} & \textbf{97.1} & 98.1 & 98.9 \\
        Amodal leaf           & 95.0 & 90.5 & 94.7 & 95.2 \\
        Amodal vein (primary) & 65.4 & 48.6 & 77.5 & 56.5 \\
        Detail vein (fine)    & 55.3 & 38.2 & 61.3 & 50.4 \\
        \bottomrule
    \end{tabular}
\end{table}

\subsection{What Makes Completion Work}
\label{sec:ablation}
\begin{figure}[t]
    \centering
    \includegraphics[width=\columnwidth]{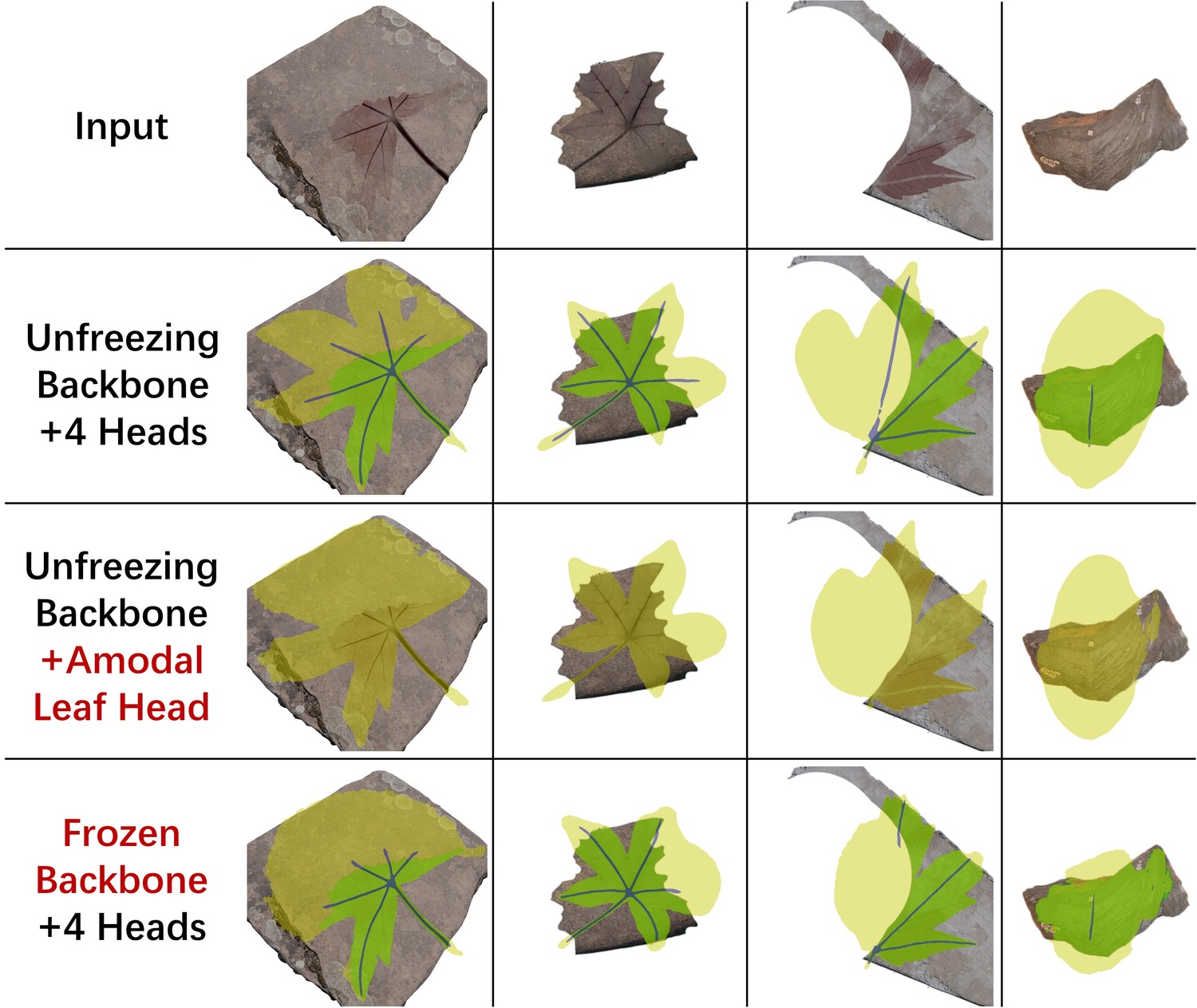}
    \caption{Ablating the two design choices. Colors follow
    Fig.~\ref{fig:teaser}, and detail vein is hidden for clarity. Row~2 is the released model; row~3 keeps the unfrozen
    backbone but trains only the amodal leaf head; row~4 keeps all four heads but
    freezes the backbone. Both ablations still complete a leaf, but they lose the lobes
    and symmetry that make the outline a leaf.}
    \label{fig:ablation}
\end{figure}

Two design choices dominate the result: unfreeze the whole backbone, and attach
all four heads. Both are simple but critical. Figure~\ref{fig:ablation} compares the full model
against dropping either choice, with resolution,
schedule, dataset, and other training parameters consistent across the three runs.

The first thing to notice is that DINOv3 is a powerful backbone, and both crippled
variants can still push a mask beyond the visible tissue into the stone or blank area.
Completion is evidently not the scarce resource here; the pretrained
representation already carries enough of a notion of ``object'' to extrapolate one.
What the two design choices buy is not \emph{whether} the model completes but
\emph{what shape} it completes into, and that is exactly the part a paleobotanist
cares about.

\begin{figure}[t]
    \centering
    \includegraphics[width=0.92\columnwidth]{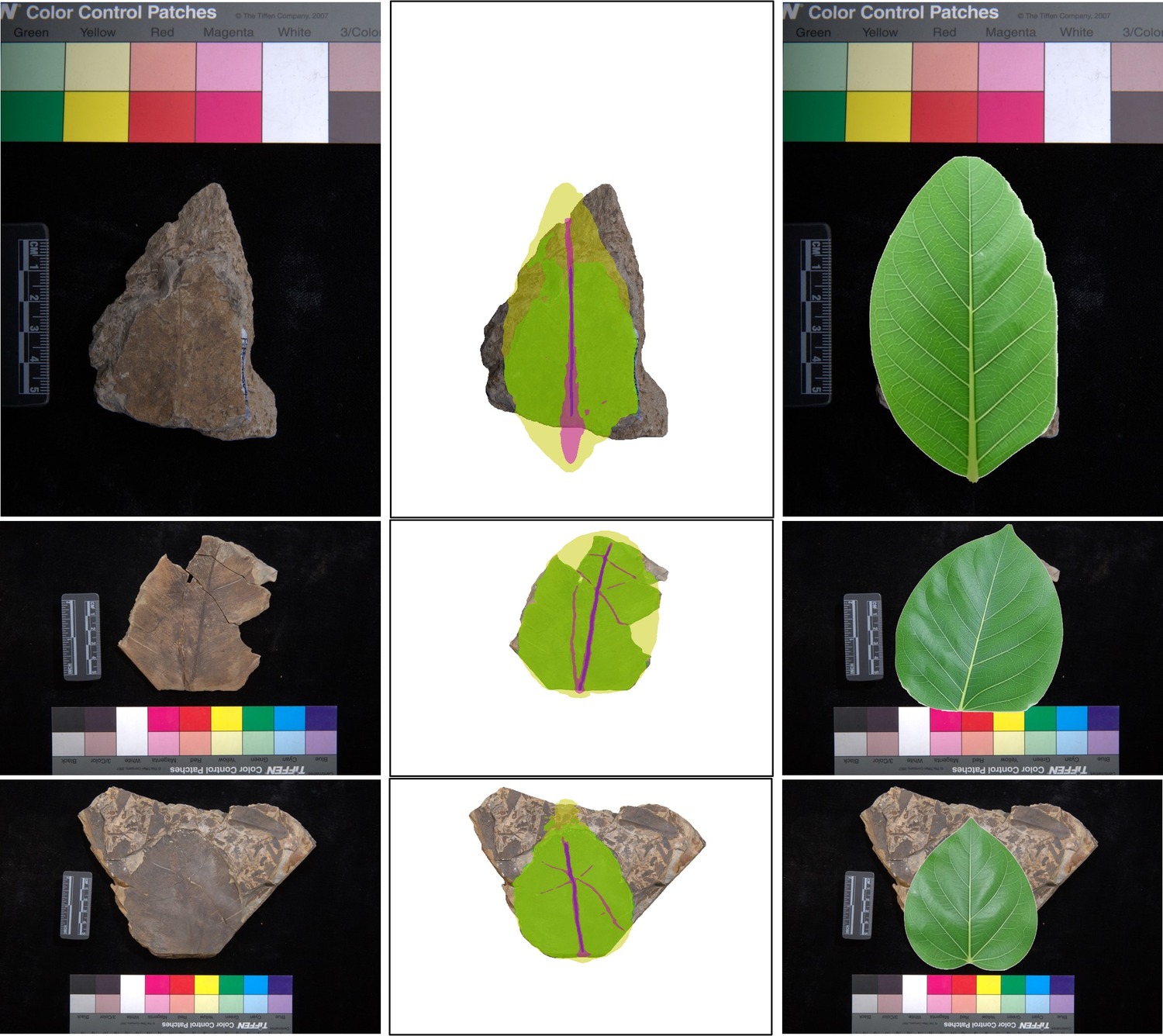}
    \caption{Generalization to real fossil specimens (photograph, predicted masks,
    optional generative rendering). None of these specimens, matrices, or capture
    setups appear in the synthetic training data. Colors follow
    Fig.~\ref{fig:teaser}.}
    \label{fig:real}
\end{figure}

\noindent\textbf{Fine-tuning versus freezing.} With the backbone frozen and only
the DPT decoder trained as recommended, the amodal head
produces a smooth, roughly elliptical envelope that is anchored on the specimen
but blind to its geometry. The pretrained features describe what is present
extremely well, but a $3{\times}3$ convolution on top of frozen tokens has no way
to reshape them into a leaf-specific shape prior. Unfreezing every layer at
$10^{-5}$ changes the character of the output within a few epochs: margins snap to
the specimen, lobes appear, and the boundary becomes
clear. The small learning rate makes this safe---the backbone
specializes to fossils without forgetting the general features that let it handle
the unseen photographs in Fig.~\ref{fig:real}.

\noindent\textbf{Head stacking.} Training the amodal head alone, with the same
unfrozen backbone, gives the second column of comparisons. Again the leaf is
completed, and the overall extent is often roughly right, but structure is
missing. In Fig.~\ref{fig:ablation} it swells past the true
margin on one side of the multi-blob leaf (column~1). Adding the visible head makes the model
say explicitly where observation ends and inference
begins. Adding the two vein heads further supplies the missing shape prior. This is
the interesting part, because a coherent venation can only be drawn on a coherent lamina,
so supervising midrib and higher-order veins forces the shared trunk to represent
lobe and sinus structure. The detail-vein head is not shown since it is often noisy,
but it still earns its place, because the two vein tasks must
agree on where real venation lies, which stops the primary-vein head from
following stone cracks. The lesson generalizes past leaves: when labels for the
target shape are scarce, an auxiliary structural task can supply an inductive bias
that direct supervision cannot.

\noindent\textbf{Capacity can hide the effect.} A note for anyone trying to
reproduce the comparison: the gap narrows as the backbone gets stronger. In earlier experiments,
with ViT-B/16 or $256^2$ resolution, the difference is more
dramatic. In other words, a larger backbone partly
compensates for a weaker training protocol, which is a good reason not to
validate such choices only at the largest scale one can afford.

\begin{table}[t]
    \centering
    \small
    \setlength{\tabcolsep}{4.5pt}
    \caption{Public amodal benchmarks (\%, higher is better). KINS \emph{test}:
    $92{,}625$ instances, $50{,}894$ occluded. COCOA-cls \emph{val}: $3{,}799$
    instances, $1{,}881$ occluded. GRASP and PLUG report the standard COCOA split
    rather than COCOA-cls, so they appear only under KINS.}
    \label{tab:public}
    \begin{tabular}{llcc}
        \toprule
        Method & Visible input & full & occ \\
        \midrule
        \multicolumn{4}{@{}l}{\emph{KINS}} \\
        VRSP~\cite{xiao2021vrsp}           & predicted mask & 80.70 & 47.33 \\
        AISFormer~\cite{tran2022aisformer} & predicted mask & 81.53 & 48.54 \\
        C2F-Seg~\cite{gao2023c2fseg}       & predicted mask & 82.22 & 53.60 \\
        GRASP~\cite{zhang2026grasp}        & predicted mask & 82.37 & 55.24 \\
        C2F-Seg~\cite{gao2023c2fseg}       & GT mask (oracle) & 87.89 & 57.60 \\
        GRASP~\cite{zhang2026grasp}        & GT mask (oracle) & \textbf{90.49} & 62.59 \\
        Amodal SAM~\cite{zhang2026amodalsam} & GT box $+$ mask & 88.79 & 63.12 \\
        PLUG (ViT-H)~\cite{liu2024plug}    & GT box prompt & 88.85 & 62.66 \\
        \textbf{Ours (DINOv3$+$DPT)}       & \textbf{none} & 85.05 & \textbf{66.65} \\
        \midrule
        \multicolumn{4}{@{}l}{\emph{COCOA-cls}} \\
        VRSP~\cite{xiao2021vrsp}           & predicted mask & 79.93 & 26.72 \\
        C2F-Seg~\cite{gao2023c2fseg}       & predicted mask & 81.71 & 36.70 \\
        Amodal SAM~\cite{zhang2026amodalsam} & GT box $+$ mask & \textbf{87.65} & \textbf{54.34} \\
        \textbf{Ours (DINOv3$+$DPT)}       & \textbf{none} & 80.90 & 38.15 \\
        \bottomrule
    \end{tabular}
\end{table}

\begin{figure}[t]
    \centering
    \includegraphics[width=\columnwidth]{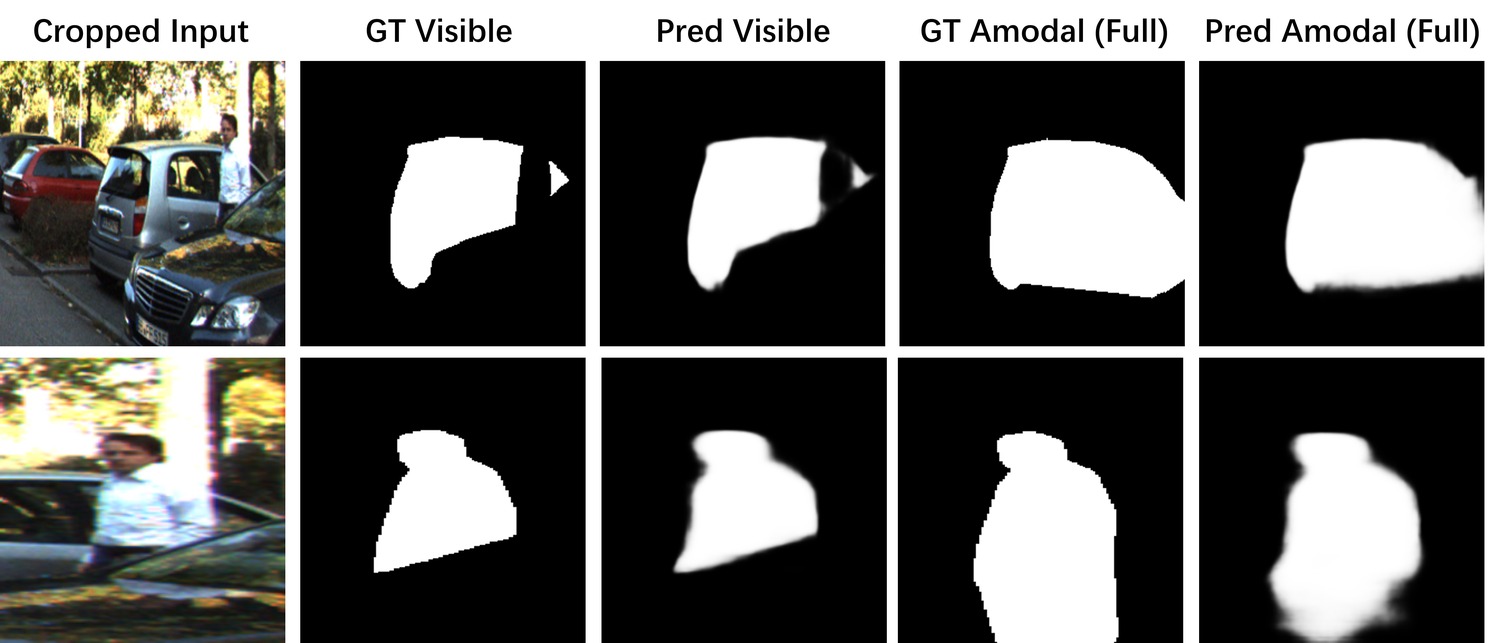}
    \caption{The two-head, RGB-only variant on occluded KINS instances. From the
    crop alone the model separates visible from occluded tissue and closes the
    car body behind the foreground vehicle (top) and the pedestrian's torso behind
    the car (bottom).}
    \label{fig:publicqual}
\end{figure}

\subsection{Amodal Benchmarks Without a Visible Mask}
\label{sec:baselines}
No public benchmark exists for fossil leaves, so we validate the
backbone/decoder choice on the standard amodal datasets. We train a
stripped-down variant---DINOv3 ViT-L/16 $+$ DPT with only the visible and amodal
heads, RGB-only, no ROI channel---separately on KINS and COCOA-cls for $30$
epochs at batch size $200$ and $256\times256$ resolution, with the same two-rate
AdamW schedule and no cross-dataset initialization. Each instance is cropped from
its ground-truth visible bounding box enlarged $2\times$ and centered in the crop,
matching the crop convention of C2F-Seg and PLUG. Evaluation is instance-level
over the full KINS test set and the COCOA-cls validation set, using full mIoU and
the occluded mIoU of Eq.~\eqref{eq:occiou}.

Centring is worth flagging, because without a visible mask it becomes part of the
task specification. What the model actually learns is ``complete the occluded
object at the centre of this crop,'' and it is the crop geometry, not an input
mask, that identifies the target. Feed it an off-centre instance and it will
happily complete whatever sits in the middle instead. This is a fair setting for
comparing against methods that crop the same way, but it is not a detector-coupled
system, and Section~\ref{sec:limitations} returns to the point.

Table~\ref{tab:public} places these numbers next to published results. Protocols
differ and we do not claim a controlled comparison; the ``visible input'' column
is the point of the table. Every baseline receives the visible region in some
form, like a predicted mask, a ground-truth mask, or at least a ground-truth box
prompt, whereas our model sees only pixels inside a crop and must decide for
itself which of them belong to the visible object.

Under this setup our model reaches $85.05$ full mIoU on KINS,
and its occluded mIoU of $66.65$ is the highest number in the table.
The pattern is what one would expect once the input protocol is
made explicit. A visible-mask input mostly buys accuracy on the visible region,
which dominates full mIoU because visible pixels outnumber occluded ones; the
occluded region instead rewards a backbone that has genuinely learned to
extrapolate, and that is what full fine-tuning provides. On COCOA-cls, which
spans $80$ diverse categories with only $\sim\!2.3$k training images, the same
variant is competitive with C2F-Seg on the occluded region but clearly behind
Amodal SAM. This is unsurprising, because our model has neither a shape prior nor a
category signal to fall back on.

\subsection{Generalization to Real Fossils}
\label{sec:realfossils}
Because all supervision is synthetic, the question that matters is whether the
model survives contact with real material. Our test set is $9$ photographs of
real fossil specimens, in which paleobotanists always put a ruler
beside the fossil. That convention
is not incidental clutter, and Section~\ref{sec:deployment} exploits it to estimate
the area of a leaf.

Figure~\ref{fig:real} shows the results. The completion remains plausible: the
model extends the blade into the rock, keeps the midrib straight, and correctly
reads a lobed leaf rather than an ellipse. Note that ground truth does not exist for these specimens,
so we treat the results as qualitative. The web deployment (Section~\ref{sec:browser}) reports
the label-free quality score of Appendix~\ref{app:quality} rather than an IoU.

\section{Deployment}
\label{sec:deployment}

A reconstruction pipeline is useful to a paleobotanist when it requires no computer vision knowledge.
The final deployment should be cross-platform and installation-free. Besides amodal segmentation, our collaborator
also wanted an estimation of the \emph{size} of leaves. And during
presentations to a general audience, they prefer the visualization of living leaves rather than a simple mask overlay.
Given these requirements, we carry out two different approaches: a browser demo and a local GPU pipeline.

\subsection{In-Browser Demo}
\label{sec:browser}
\begin{figure}[t]
    \centering
    \includegraphics[width=\columnwidth]{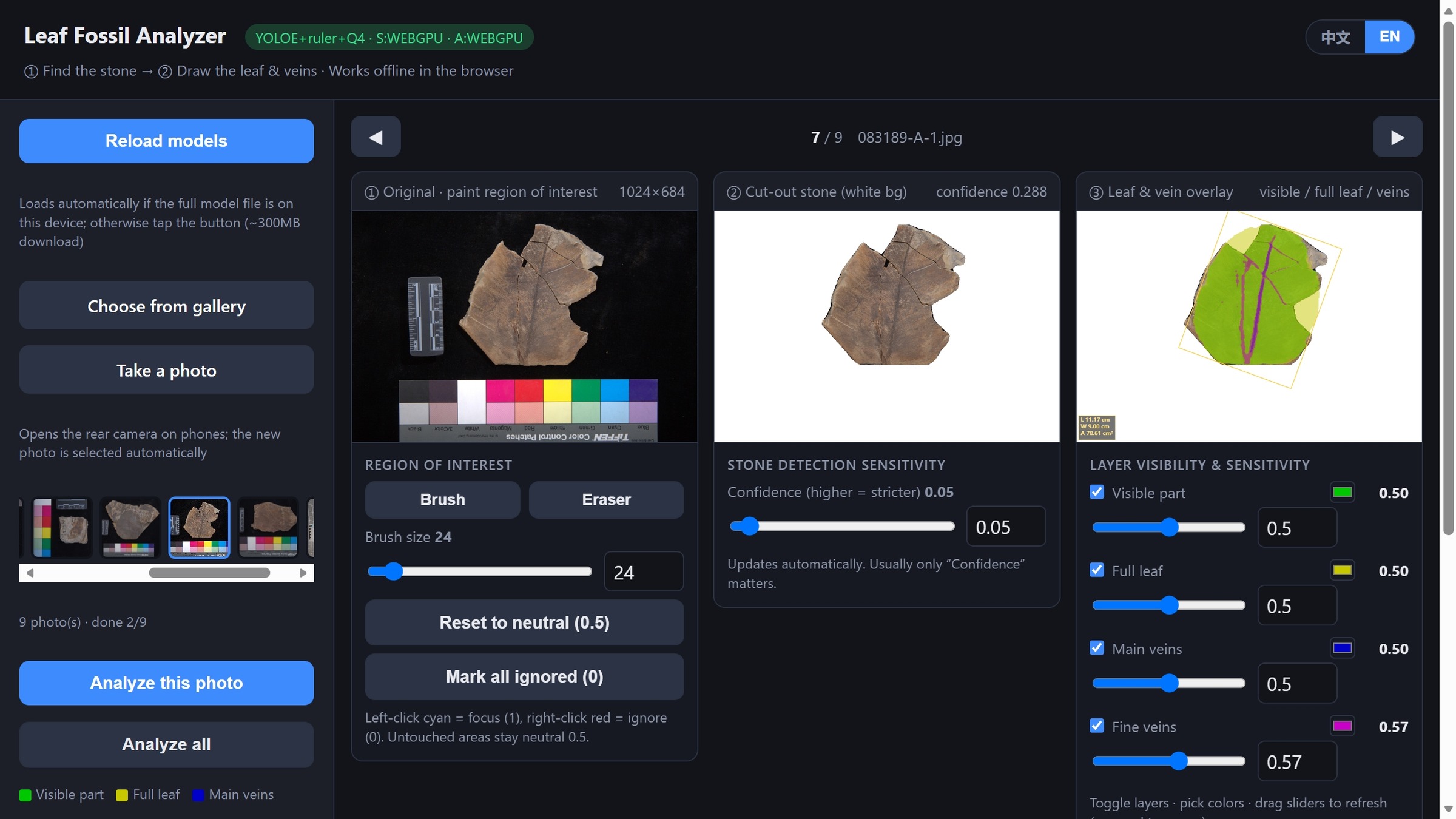}
    \caption{The offline browser demo, running locally in WebGPU on the 4-bit
    model. Panel~1: the photograph with the ROI brush, where painting writes $1$,
    erasing writes $0$, and untouched pixels stay neutral at $0.5$. Panel~2: the
    YOLO26-seg stone cut-out on white. Panel~3: the four masks as toggleable overlays
    with live per-head thresholds, plus the ruler-derived size box.}
    \label{fig:browser}
\end{figure}

The browser demo (Fig.~\ref{fig:browser}) runs entirely client-side over a static
HTTP server and performs no network inference. It chains two models.

\noindent\textbf{Stone cutout.} A real photo contains the stone specimen, the
background, and often a ruler and a color chart. We first run
YOLO26-seg~\cite{wang2025yoloe}, a real-time open-vocabulary detector-segmenter, with
the prompt \texttt{stone} (and ruler prompts, below), and composite the top stone
instance onto a white background to match the training distribution. If no stone
is detected the pipeline skips completion rather than hallucinating from an empty
input.

\noindent\textbf{Amodal inference.} \method is exported to ONNX and quantized to
4-bit weights (\texttt{MatMulNBits}), reducing the model from $\approx$1.27\,GB
in fp32 to $\approx$269\,MB---small enough to download once and cache in a
browser. It runs through ONNX Runtime Web~\cite{onnxruntime} on the WebGPU
backend. If anything goes wrong, it falls back to the WASM backend.

\noindent\textbf{Quantization fidelity.} Weight-only 4-bit quantization is
usually justified on language benchmarks, so we measured it directly on the task.
Table~\ref{tab:quant} compares the browser model against the fp32 PyTorch
checkpoint on nine real fossil photographs, treating the fp32 prediction as
reference. The amodal leaf agrees at $0.910$ IoU and the primary vein at $0.845$,
with soft-probability correlations above $0.99$. Disagreement is concentrated at
mask boundaries, which is why IoU drops more than correlation. In terms of qualitative
perception, the 4-bit model is still acceptable.
\begin{table}[t]
    \centering
    \small
    \caption{4-bit browser model versus the fp32 PyTorch checkpoint on nine real
    fossil photographs, with fp32 treated as reference.}
    \label{tab:quant}
    \begin{tabular}{lcccc}
        \toprule
        Head & IoU & F1 & prob.\ corr.\ & prob.\ MAE \\
        \midrule
        Amodal leaf          & 0.910 & 0.951 & 0.997 & 0.014 \\
        Amodal vein          & 0.845 & 0.915 & 0.991 & 0.001 \\
        \bottomrule
    \end{tabular}
\end{table}

\noindent\textbf{Interaction.} The four heads are shown as colored overlays with
independent, live-adjustable thresholds, colors, and on/off toggles. The user
can brush an ROI where painting writes $1$, erasing writes $0$, and untouched pixels
remain at the neutral $0.5$. Since the model itself is already good enough, we
do not use ROI very often.

\noindent\textbf{Physical scale and size.} Paleobotanists take photos with
a ruler as a matter of routine, so the calibration information we need is
already in the image. YOLO26-seg detects the ruler via several
natural-language prompts (\eg \texttt{ruler}, \texttt{scale bar}, \texttt{cm
scale}); the demo estimates pixels-per-centimetre from the detected scale bar
and calculates the length, width of the bounding box and the area of amodal leaf mask.

\noindent\textbf{No-ground-truth quality.} Because real fossils have no labels,
we compute a self-contained quality score for each head so a user can judge a
prediction without ground truth. It combines a confidence component (mean
predicted probability inside versus outside the mask, their margin, and the
separation of both from the decision threshold) with a shape component
(compactness, boundary smoothness, and axis-aligned symmetry of the largest
connected component), as $Q=0.65\,Q_{\text{conf}}+0.35\,Q_{\text{shape}}$;
Appendix~\ref{app:quality} gives the formulas. It is a plausibility check, not an
accuracy estimate, so we mainly use it to flag unsure inputs.

\subsection{Local GPU Pipeline}
\label{sec:localgpu}
For users with an NVIDIA GPU, we provide a Python pipeline that trades the
browser's zero-install convenience for higher-quality segmentation and an
optional generative step. Stone and ruler cutout is performed by a compact
promptable segmenter, an EfficientSAM3 variant with a TinyViT image backbone
and a MobileCLIP~\cite{vasu2024mobileclip} text encoder---driven by the same text
prompts as the browser, with text embeddings cached so repeated runs are fast.
\method then runs in autocast bfloat16, with weights stored on disk in fp16 format.
Inference takes $45$\,ms per image on RTX4080. The pipeline outputs the
white-background stone, the per-head masks, an overlay with the calibrated size
box, and a table of measurements for batch processing.

\noindent\textbf{Optional leaf revival.} Amodal masks turned out to be less attractive for the general public,
while the reconstructed living leaves drew the strongest reaction in informal demonstrations.
So, we added it as a final optional stage. The
pipeline conditions a Flux.2 diffusion model~\cite{blackforest2025flux2} on a text
prompt plus the predicted amodal leaf and amodal vein masks, optionally with the
cropped fossil or a photograph of a related extant leaf as an appearance
reference, and it repaints a \emph{living} leaf filling the reconstructed
outline (right column of Fig.~\ref{fig:teaser}). Feeding both masks matters: the
leaf mask fixes the silhouette while the vein mask keeps the generated venation
aligned with what the model actually inferred, rather than letting the diffusion
prior invent its own. We choose Flux.2-Klein because it is the smallest
quantized release of the latest open-weight diffusion family which can run on a single RTX4080.
This stage is only for visualization now. In the future, we hope to improve the
segmentation result of detail veins so that it can be used to condition the diffusion model better.

\section{Limitations}
\label{sec:limitations}

\begin{figure}[t]
    \centering
    \includegraphics[width=0.95\columnwidth]{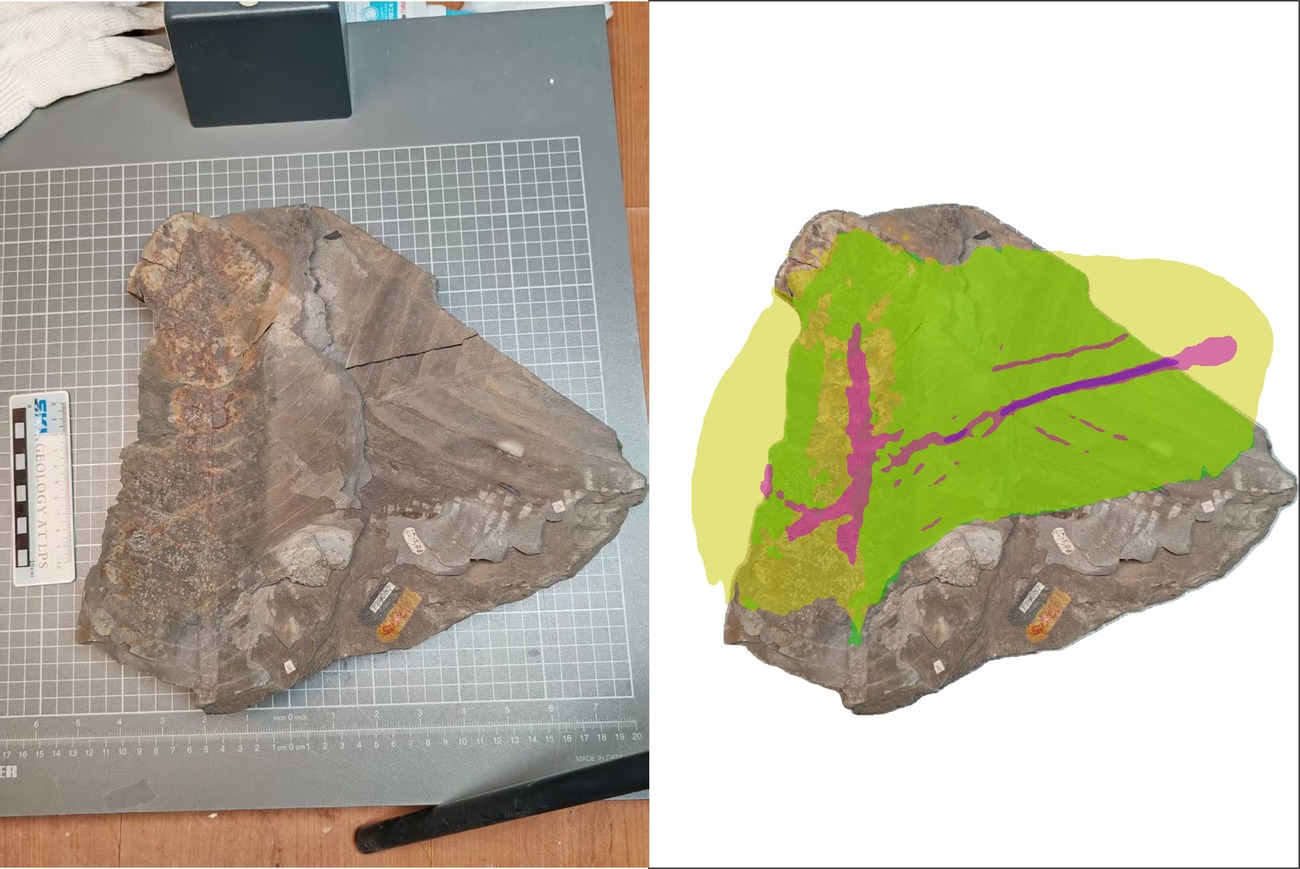}
    \caption{Failure mode: two leaf fossils on one slab. The amodal head (pale
    yellow) merges both specimens into a single silhouette, since the model has no
    instance mechanism. The vein head (magenta) nonetheless recovers two distinct
    midribs with different orientations, indicating the backbone does separate the
    two leaves and that the limitation lies in the output parameterization.}
    \label{fig:failure}
\end{figure}

\noindent\textbf{One leaf per image.} The most consequential limitation is
structural: \method predicts four masks for an image, so it implicitly assumes the
photograph contains a single leaf fossil. It performs amodal \emph{semantic}
segmentation, not amodal \emph{instance} segmentation, and it has no mechanism for
saying ``there are two leaves here.'' Figure~\ref{fig:failure} shows what happens
on a slab bearing two overlapping specimens: the amodal head merges them into one
oversized silhouette. But thanks to the generalizing ability of DINOv3,
the two separate midribs are recognized, even though nothing similar exists in the training data.
Further improvements includes query- or prompt-based heads in the spirit of instance
segmentation, and a multi-leaf dataset for training and validation. The same limitation appears in the public
benchmarks from the other direction, where target identity is carried by the crop
being centered (Section~\ref{sec:baselines}) rather than by any input mask.

\noindent\textbf{Other limitations.} Training is fully synthetic, so there remains a domain gap
to real specimens. Fine venation under heavy occlusion is unreliable, and detailed
characteristics like the tip of tropical leaves cannot be recovered. The fixed $448^2$ working resolution compresses
original information, and the ruler detector is specially designed for two typical rulers.
Finally, the generative visualization can hallucinate details so it is only for illustration now.

\section{Conclusion}
\label{sec:conclusion}
We framed leaf-fossil analysis as amodal reconstruction and presented \method, a
fully fine-tuned DINOv3\,$+$\,DPT model with four specialized heads, which recovers
complete leaf shape and venation from a single photograph without ever being told
which pixels are visible. It reaches $95.0\%$ Dice on the amodal leaf and transfers
to real fossil specimens. The two changes that carry the result are remarkably simple: 
unfreeze the whole backbone at a small learning rate, and
attach auxiliary venation heads. These findings are worth carrying
to other amodal-related problems.

Stripped to two heads and
RGB input, the same recipe reports the highest occluded mIoU on KINS without visible mask
input, suggesting that the
finding is not specific to leaves. The clearest next step is instance-level
prediction, so that a slab with several specimens yields several leaves. Beyond
that, we also look forward to real annotated fossils, detector-coupled evaluation without ground-truth
crops, stronger multi-scale vein modelling, and closing the synthetic-to-real
domain gap.

\bibliographystyle{IEEEtran}
\bibliography{references}

\clearpage
\appendices

\section{Full Implementation Details}
\label{app:implementation}

Table~\ref{tab:hparams} lists every hyperparameter used to train the released
checkpoint \texttt{amodal\_\brk dino\_\brk vitl16\_\brk roi\_\brk full\_\brk
indep.pt}. The reference implementation lives in
\texttt{Leaf\_\brk Completion\_\brk DINO/}, split across \texttt{encoder.py},
\texttt{decoder.py}, \texttt{model.py}, \texttt{roi\_\brk patch\_\brk embed.py},
\texttt{heads.py}, \texttt{losses.py} and \texttt{data.py}, with
\texttt{train.py} as the entry point.

\begin{table}[h]
    \centering
    \footnotesize
    \setlength{\tabcolsep}{4pt}
    \caption{Complete training configuration for the released fossil model.}
    \label{tab:hparams}
    \begin{tabular}{@{}>{\raggedright\arraybackslash}p{0.38\columnwidth}
                      >{\raggedright\arraybackslash}p{0.56\columnwidth}@{}}
        \toprule
        Item & Setting \\
        \midrule
        Backbone & DINOv3 ViT-L/16, LVD-1689M weights \\
        Dim / patch / blocks & 1024 / 16 / 24 \\
        Feature blocks & $\{4,11,17,23\}$ (four evenly spaced) \\
        Decoder & official DPT trunk, \texttt{readout=project} \\
        Trunk width / post-proc & 256 / $[128,256,512,768]$ \\
        Heads & 4 independent $3{\times}3$ conv ($256{\to}1$) \\
        ROI injection & 4-channel patch embed, RGB-mean init \\
        ROI encoding & $0.5$ neutral / $1.0$ hint; normalized with mean and std $0.5$ \\
        Input & RGB(3)+ROI(1), $448{\times}448$ \\
        Backbone mode & full fine-tuning (unfrozen) \\
        Optimizer & AdamW, weight decay $10^{-4}$ \\
        LR (trunk / heads) & $10^{-4}$ \\
        LR (backbone) & $10^{-5}$ \\
        Schedule & cosine to $10^{-6}$ \\
        Epochs & 40 ($\approx$123\,min total) \\
        Batch size & 64 \\
        Precision & bfloat16 autocast, channels-last \\
        Gradient clipping & max-norm $1.0$ \\
        Hardware & 1$\times$ NVIDIA RTX PRO 6000 (Blackwell) \\
        Model selection & best val amodal Dice (epoch 37) \\
        \midrule
        \multicolumn{2}{@{}l}{\emph{Loss weights}} \\
        Head weights $\alpha$ & $1.0 / 1.0 / 1.0 / 0.4$ (v / a / av / dv) \\
        Completion reweight $\lambda_c$ & $2.0$ \\
        Tversky $\beta_{\text{FP}},\beta_{\text{FN}}$ & $0.65,\;0.35$ \\
        Area overflow (a / av) & $0.5$ / $0.5$, power $2$ \\
        FP-pixel penalty & $0.25$ \\
        Containment $\gamma$ & $0.5$ (v$\subseteq$a), $0.5$ (av$\subseteq$a), $0.3$ (dv$\subseteq$a) \\
        Boundary BCE weight & $1+2T+4\,\partial T$ \\
        \midrule
        \multicolumn{2}{@{}l}{\emph{Augmentation}} \\
        Flips & horizontal and vertical, $p=0.5$ each \\
        Rotation & $k\cdot 90^{\circ}$, $k\sim\mathcal{U}\{0,1,2,3\}$ \\
        Scale & $\mathcal{U}(0.5,1)$ about a random point, pad white \\
        Color jitter & brightness $[0.75,1.65]$, contrast $0.15$, saturation $0.1$, hue $0.05$ \\
        ROI source & uniform over \{stone, visible, amodal, amodal-vein, neutral\} \\
        ROI corruption & dilate/erode $k\in\{3,5,7,9\}$, $1$--$3$ iterations; elastic warp $p=0.9$, $\alpha\sim\mathcal{U}(6,22)$, $\sigma\sim\mathcal{U}(3,7)$ \\
        \bottomrule
    \end{tabular}
\end{table}

\section{The ROI Hint}
\label{app:roi}

The ROI channel is an interface feature, not part of the core method: every fossil
number in the paper is measured with a neutral hint, and the public-benchmark
variant omits the channel entirely. We designed it at an early stage, but eventually
the model becomes good enough to complete without any hint.

\noindent\textbf{Encoding.} A binary hint invites the encoding $\{0,1\}$, but that
conflates two different statements: ``this region is irrelevant'' and ``no hint was
given.'' In deployment the second is the common case---users usually paint
nothing---so encoding it as $0$ would feed a strong constant signal into the stem on
almost every query. We instead use $1$ for a positive hint, $0.5$ for no
information, and normalize the channel with mean and standard deviation $0.5$, so a
neutral hint maps exactly to zero and contributes nothing to the patch embedding.
The value $0$ never occurs during training but remains available at inference as an
out-of-distribution ``suppress'' stroke, which in practice pushes completion away
from the painted region; the browser exposes it as the eraser.

\noindent\textbf{Injection point.} The hint enters at the stem, through the widened
4-channel patch-embedding convolution. We also implemented the obvious
alternative---keep the backbone RGB-only and add a $1{\times}1$ projection of the
hint to the DPT trunk features. In-backbone injection is preferable because
self-attention can reason about the hint from the first layer rather than after the
representation is already formed, and it is the default for all reported results.

\noindent\textbf{Sampling and corruption.} The hint source is drawn uniformly per
sample from $\{$stone, visible, amodal, amodal-vein, neutral$\}$, so one set of
weights handles a stone cut-out, a tight visible region, an oracle amodal region, a
vein scribble, and no hint at all. The chosen mask is then corrupted by random
dilation or erosion (kernel $\in\{3,5,7,9\}$, one to three iterations) and, with
probability $0.9$, an elastic warp ($\alpha\sim\mathcal{U}(6,22)$,
$\sigma\sim\mathcal{U}(3,7)$). Corruption is the load-bearing part. Without it the
model treats the hint as ground truth, learns to trace its boundary, and falls apart
the moment a user draws a sloppy stroke. Vein hints are only dilated, never eroded,
since a one-pixel structure does not survive erosion. At validation the hint is the
stone mask where available and neutral otherwise.

\section{Data and Reproducibility Notes}
\label{app:data}

\noindent\textbf{Pipeline summary.} $160$ leaves hand-picked from the $4{,}041$
NMNS cleared-leaf photographs, segmented with SAM~3 and vein-labelled with OpenCV
plus manual correction $\rightarrow$ $2{,}200$ Blender composites at
$1024\times1024$ ($2{,}000$ train scenes, $200$ val scenes) $\rightarrow$ four
OpenCV breakage variants per composite $\rightarrow$ $10{,}000$ train $/$ $1{,}000$
val images. The split is by scene, so no leaf instance and no stone appears on both
sides. The test set is disjoint in kind rather than in split: $9$ photographs of
real specimens, never used for training or model selection.

\noindent\textbf{Directory layout.} \texttt{leaf\_fossil/\{train,val\}/} with
parallel subfolders \texttt{leaf\_fossil}, \texttt{visible\_mask},
\texttt{amodal\_mask}, \texttt{vein\_mask}, \texttt{detail\_vein\_mask}, and
\texttt{stone\_mask}. Filename prefixes $\{$\texttt{(none)}, \texttt{broken\_},
\texttt{broken2\_}, \texttt{broken3\_}, \texttt{broken4\_}$\}$ mark the five damage
levels of a scene. Because amodal and vein masks are stored once per scene, the
loader resolves a mask by exact basename and otherwise strips leading
\texttt{prefix\_} tokens one at a time. Images are resized to the working
resolution with bilinear interpolation, masks with nearest-neighbour interpolation
followed by re-binarization at $127$, and the whole decoded split is held in RAM.

\begin{figure}[h]
    \centering
    \includegraphics[width=\columnwidth]{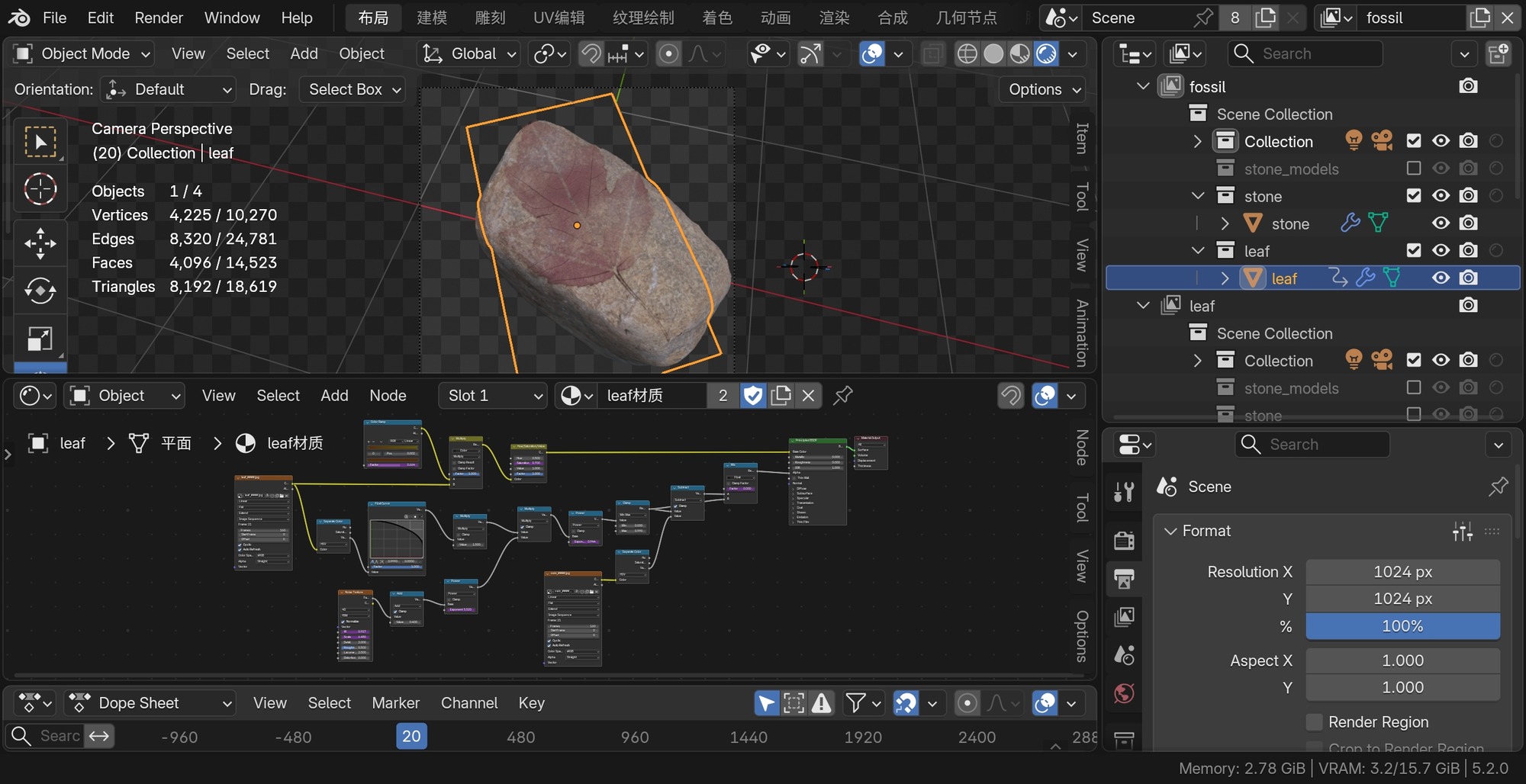}
    \caption{The Blender compositing scene. A segmented cleared-leaf texture is
    projected onto a randomized stone mesh; the shader graph controls wear and
    staining of the lamina, and camera, lighting, and pose are randomized per
    frame. Gross breakage is applied afterwards in OpenCV.}
    \label{fig:blender}
\end{figure}

Figure~\ref{fig:blender} shows the Blender scene used for compositing. The leaf
plane, the stone mesh, and the shader graph that mixes noise textures into the
leaf material are all driven by seeded random parameters, and each render pass
writes one RGB frame and its five masks.

\noindent\textbf{Public benchmark variant.} The numbers in
Table~\ref{tab:public} come from a separate two-head model
(visible and amodal only, RGB input, no ROI channel) trained per dataset: DINOv3
ViT-L/16 with full fine-tuning, $256\times256$ crops from the ground-truth
visible box enlarged $2\times$, batch size $200$, $30$ epochs, cosine schedule,
best checkpoint by validation full mIoU (epoch $26$ on KINS, epoch $22$ on
COCOA-cls), no initialization from the other dataset. KINS masks are decoded from
polygons and COCOA-cls masks from RLE. We use the COCOA-cls \texttt{with\_classes}
split ($\sim$2{,}276 train / 1{,}223 val images); some papers report a slightly
larger train split, which should be kept in mind when reading the COCOA-cls
block. During training we validate on a fixed $10$k-instance KINS
subset, which gave $85.05$ / $66.65$---within $0.1$ of the full test set, so the
subset was a faithful proxy.

\section{No-Ground-Truth Quality Score}
\label{app:quality}

For deployment on unlabeled real fossils we score each head from its own soft
output. Let $p_i=\sigma(y_i)$ be per-pixel probabilities, and let the binary mask
be the largest connected component of $\{p_i>\tau\}$ with $\tau=0.5$. Define the
mean probabilities inside and outside the mask, $\bar{p}_{\text{in}}$ and
$\bar{p}_{\text{out}}$, their margin $\bar{p}_{\text{in}}-\bar{p}_{\text{out}}$,
and a threshold-separation term that rewards $\bar{p}_{\text{in}}$ lying above
and $\bar{p}_{\text{out}}$ below $\tau$. The confidence score $Q_{\text{conf}}$
is a fixed convex combination of these four quantities with weights
$(0.35, 0.25, 0.20, 0.20)$. The shape score $Q_{\text{shape}}$ combines
isoperimetric compactness $4\pi A/P^2$, a boundary-smoothness proxy, and
axis-aligned symmetry (the larger of the horizontal- and vertical-flip IoU of the
mask) with weights $(0.4, 0.3, 0.3)$. The reported quality is
\begin{equation}
Q = 0.65\,Q_{\text{conf}} + 0.35\,Q_{\text{shape}}.
\end{equation}

On the nine real fossils of Table~\ref{tab:quant} the fp32 model averages
$Q=0.818$ for the amodal leaf ($Q_{\text{conf}}=0.942$,
$Q_{\text{shape}}=0.589$) and $Q=0.697$ for the primary vein. The shape term is
structurally low for veins---a one-pixel curve has near-zero isoperimetric
compactness---so vein scores should be compared against other veins rather than
against leaves. The identical computation is implemented in the browser
(\texttt{amodal\_\brk quality.js}) and in the local pipeline
(\texttt{quality\_\brk metrics.py}) so that on-device scores match.

\section{Deployment Specifications}
\label{app:deploy}

\noindent\textbf{Browser.} Stone and ruler detection uses a YOLO26-seg model
exported to ONNX ($\approx$40\,MB) at $640\times640$ letterboxed input with
$160\times160$ mask prototypes and $32$ mask coefficients, plus a stone-only
checkpoint as fallback when no ruler is present. \method is exported as a 4-bit
(\texttt{MatMulNBits}) ONNX model ($\approx$269\,MB, split into $<$100\,MB shards
for static hosting) with input \texttt{rgb\_\brk roi} of shape $[1,4,448,448]$
and output \texttt{logits} of shape $[4,448,448]$; it is served through ONNX
Runtime Web (WebGPU, WASM fallback) with the output-sanity check described in
Section~\ref{sec:browser}. The full offline bundle is $\approx$390\,MB and requires only a static HTTP server
and a WebGPU-capable Chromium browser.

\noindent\textbf{Local GPU.} Stone and ruler cutout uses an EfficientSAM3-style
promptable segmenter (TinyViT-11M image backbone, MobileCLIP-S0 text encoder,
context length $16$) in bfloat16 with cached text embeddings. \method weights are
stored in fp16 (overflow-sensitive layers in bfloat16) and cast to fp32 at load,
with inference under bfloat16 autocast. The bundle is $\approx$0.9--1.0\,GB
(segmenter $\approx$0.2\,GB, amodal fp16 $\approx$0.65\,GB) and requires a
CUDA-enabled PyTorch. The optional generative leaf-revival stage loads an fp8
Flux.2-Klein model on demand and is off by default.

\end{document}